\documentclass[conference]{IEEEtran}
\IEEEoverridecommandlockouts
\usepackage{cite}
\usepackage{amsmath,amssymb,amsfonts}
\usepackage{algorithm}
\usepackage{algpseudocode}

\usepackage{placeins}
\usepackage{booktabs}
\usepackage{multirow}
\usepackage{adjustbox}

\makeatletter
\renewcommand{\fps@algorithm}{htbp}
\makeatother

\usepackage{graphicx}
\usepackage{textcomp}
\usepackage{xcolor}
\usepackage{url}
\def\BibTeX{{\rm B\kern-.05em{\sc i\kern-.025em b}\kern-.08em
    T\kern-.1667em\lower.7ex\hbox{E}\kern-.125emX}}
\begin{document}

\title{A Grounded and Decomposed Framework for Relation-Level Hallucination Evaluation in Abstractive Summarization}

\author{\IEEEauthorblockN{1\textsuperscript{st} Praveen Kumar Katwe}
\IEEEauthorblockA{\textit{Dept. of CSE} \\
\textit{IIIT Bhubaneswar}\\
Bhubaneswar, Odisha, India \\
c121007@iiit-bh.ac.in}
\and
\IEEEauthorblockN{2\textsuperscript{nd} Rakesh Chandra Balabantaray}
\IEEEauthorblockA{\textit{Dept. of CSE} \\
\textit{IIIT Bhubaneswar}\\
Bhubaneswar, Odisha, India \\
rakesh@iiit-bh.ac.in}
\and
\IEEEauthorblockN{3\textsuperscript{rd} Kali Prasad Vittala}
\IEEEauthorblockA{\textit{Informatica} \\
\textit{}\\
Bengaluru, Karnataka, India \\
kprasad@informatica.com}
\and
\IEEEauthorblockN{4\textsuperscript{th} Naman Kabadi}
\IEEEauthorblockA{\textit{TEKsystems Global Services} \\
\textit{}\\
Bengaluru, Karnataka, India \\
namankabadi50@gmail.com}
}

\maketitle

\begin{abstract}
Abstractive text summarization systems frequently generate fluent yet unfaithful summaries by fabricating or distorting relationships between entities and events. Such relation-level hallucinations undermine the reliability of generated summaries, particularly in high-stakes domains. In this work, we present a refined and grounded framework for evaluating relation hallucination in abstractive summarization. 

We present the empirical Relation Hallucination Index (RHI) by introducing a dependency-aware relation extraction algorithm that incorporates lemmatization-based normalization, named entity grounded subject resolution, passive agent recovery, negation-aware verb modeling, reporting verb filtering, nominal relation fallback, clausal propagation, and systematic deduplication. These enhancements improve the structural fidelity of extracted relation triples and reduce spurious matches during evaluation.

In addition, we introduce a normalized formulation of RHI to ensure scale-invariant comparison between datasets and models. The revised metric decomposes hallucination into interpretable components, aggregates relation hallucination metric into a normalized relation faithfulness score.

Extensive evaluation across multiple state-of-the-art summarization models demonstrates that the grounded extraction process yields more stable and discriminative hallucination measurements.  The proposed framework advances automated relation-level faithfulness evaluation and supports coherence-aware, hallucination-sensitive model analysis.
\end{abstract}


\section{Introduction}

Recent advances in natural text generation have significantly improved the fluency and readability of automatically generated summaries. However, fluency does not guarantee factual reliability. Abstractive summarization systems frequently introduce relational distortions, where entities are preserved but the connections between them are inaccurately constructed. Such relation-level hallucinations can subtly alter the meaning of source content, raising serious concerns in applications where correctness is critical.

Although existing evaluation metrics emphasize lexical overlap or entity matching, they often fail to capture inconsistencies in structured relationships. Assessing relational faithfulness therefore requires explicit modeling of subject–verb–object structures and systematic comparison between source documents, references, and generated summaries.

In this work, we present a grounded and normalized framework for evaluating relation-level hallucination. We refine relation extraction using linguistically informed mechanisms and introduce a normalized formulation of the Relation Hallucination Index (RHI) to enable stable comparison across models and datasets. Our approach provides a structured and interpretable mechanism for analyzing relational fidelity and coherence \cite{Reimers2019SBERT} in abstractive summarization systems.

\section{Background and Problem Formulation}

\subsection{Relation-Level Hallucination}

Relation-level hallucination arises when a summarization system preserves surface entities from the source document, but incorrectly constructs or alters the relationships among them. Modern transformer-based models and large language models generate summaries by predicting tokens conditioned on contextual representations rather than explicitly modeling structured relational dependencies. As a result, these systems may inadvertently introduce fabricated interactions, misattribute actions, or exaggerate causal links between entities. Unlike simple factual omissions, relational distortions modify the semantic structure of the original content and may lead to misleading interpretations. Detecting such inconsistencies, therefore, requires a structured comparison of subject–predicate–object relations across source texts, references, and generated summaries. So, relation-aware extraction and evaluation frameworks are essential to quantify these structural deviations and assess summary faithfulness beyond lexical similarity.

\subsection{Limitations of Existing Metrics}

Widely adopted evaluation measures such as ROUGE primarily rely on n-gram overlap, which captures surface similarity but does not reflect relational correctness. Embedding-based metrics, including BERT-derived similarity scores, assess semantic proximity yet lack explicit modeling of structured entity interactions. Entity-centric metrics improve entity coverage evaluation but remain insensitive to incorrect predicate assignments or distorted connections between entities. Consequently, existing approaches often fail to identify relation-level inconsistencies, motivating the need for structured and normalized relational evaluation mechanisms.

\section{Related Work}

\subsection{Evaluation Metrics for Summarization}

Automatic evaluation of summarization systems has historically relied on lexical overlap measures, most prominently ROUGE  \cite{Lin2004ROUGE}. These metrics quantify n-gram correspondence between system outputs and reference summaries and remain widely adopted for benchmarking informativeness. However, lexical similarity does not guarantee factual correctness, and high overlap scores may still accompany structurally distorted content. As neural generation models became more expressive, limitations of purely surface-based metrics became increasingly apparent, motivating research into faithfulness-oriented evaluation strategies.

\subsection{Approaches to Mitigating Hallucination}

To improve the reliability of abstractive summaries, several modeling strategies have been proposed. Entity-aware decoding mechanisms incorporate named entity signals during generation to encourage preservation of salient source entities \cite{zhou2021entity}. Knowledge-grounded frameworks further attempt to constrain generation by referencing structured repositories such as knowledge graphs \cite{chen2020kgpt}. Other approaches employ adversarial training, where a discriminator guides the generator toward producing outputs that are not only fluent but also factually aligned with source content \cite{Wu2022Precisely}. While these methods enhance generation quality, they focus primarily on model training rather than post-hoc structural evaluation.

\subsection{Entity-Level Faithfulness Metrics}

Beyond generation strategies, evaluation metrics specifically targeting hallucination have been introduced. Entity-level measures, such as Entity F1 \cite{lample2016neural}, compare named entities across source and generated summaries to quantify factual preservation. Subsequent work expanded this perspective by categorizing hallucinations into subject, object, and relation dimensions \cite{Maynez2020Faithfulness}. Entity-centric indices, including the Entity Hallucination Index (EHI) \cite{Praveenkumar2023EHI}, provide finer-grained insight into entity consistency. Nevertheless, entity-level comparison does not guarantee relational correctness. A summary may preserve correct entities while misassigning roles, altering predicates, or fabricating interactions between otherwise accurate mentions.

\subsection{Structured and Knowledge-Based Evaluation}

Structured evaluation methods model textual content as relational triples or knowledge graph representations. Benchmarks such as Text2KG frameworks \cite{mihindukulasooriya2023text2kgbench} evaluate extraction quality under predefined ontologies. While effective for structured prediction tasks, such approaches typically require domain-specific schemas or world knowledge alignment, limiting their applicability to general summarization evaluation. Moreover, many existing relation-based pipelines rely on basic extraction heuristics that are sensitive to syntactic variation and duplication artifacts, potentially affecting hallucination measurements.

\subsection{Gap and Motivation}

Existing research demonstrates substantial progress in entity preservation and semantic similarity evaluation; however, systematic quantification of relation-level hallucination without reliance on external ontologies remains underexplored. In particular, prior methods lack linguistically grounded extraction refinements and normalized scoring mechanisms that ensure stable cross-dataset comparison. The present work addresses these gaps by integrating dependency-aware relation extraction, refined hallucination decomposition, normalized RHI computation, and coherence-aware relational continuity analysis within a unified evaluation framework.

\section{Methodology}

The proposed framework consists of four major components: dataset preparation, summary generation, relation extraction, and relation-level hallucination evaluation. The overall pipeline constructs structured relational representations from input, reference, and generated summaries, and evaluates relational faithfulness using a normalized Relation Hallucination Index.

\subsection{Dataset Selection and Preprocessing}

\subsubsection{Dataset Selection}

To evaluate relation-level hallucination across varying abstraction styles, we employ three widely used summarization benchmarks covering complementary news domains. The experimental dataset consists of \textbf{XSUM (200 instances)}, \textbf{XLSUM (200 instances)}, and the \textbf{CNN/DailyMail SumEval subset (400 instances)}, resulting in a total of \textbf{800 evaluation samples}.

XSUM is selected for its highly abstractive single-sentence summaries, which frequently induce factual distortions. XLSUM introduces stylistic and linguistic diversity through cross-domain news reporting, enabling robustness assessment under varied writing structures. The CNN/DailyMail SumEval dataset is incorporated due to its established use in factual consistency evaluation (e.g., SummaC), providing longer multi-sentence summaries with stronger extractive grounding. 

This combination enables balanced evaluation across extreme abstraction, moderate abstraction, and factual consistency-oriented summarization settings, improving generalizability of relation hallucination analysis.

\subsubsection{Dataset Preparation}

\begin{figure}[t]
\centering
\includegraphics[width=0.45\textwidth]{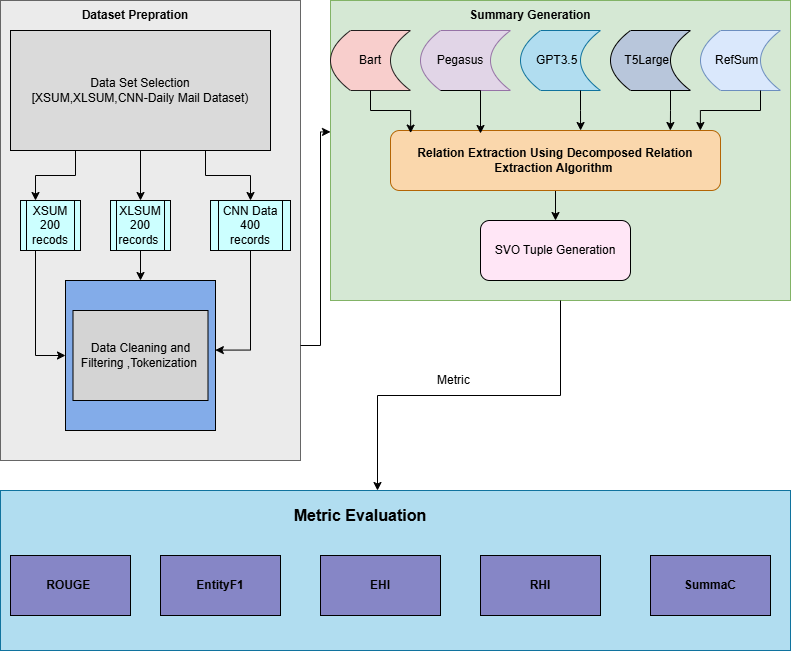}
\caption{Process flow diagram illustrating dataset preparation, summary generation, relation extraction, and metric evaluation pipeline.}
\label{fig:process_flow}
\end{figure}

All documents undergo normalization, removal of noisy symbols, and sentence-level segmentation prior to processing. Inputs exceeding transformer token limits are truncated while preserving contextual coherence. The cleaned texts are subsequently passed to the relation extraction pipeline for Subject--Verb--Object (SVO) tuple generation, as illustrated in Fig.~\ref{fig:process_flow}. 

\subsection{Summary Generation}

\subsubsection{Model Selection}

We evaluate four representative transformer-based abstractive summarization models, namely \textbf{BART-large-CNN}, \textbf{PEGASUS}, \textbf{T5-large}, and \textbf{GPT-3.5}, covering diverse pretraining objectives and abstraction behaviors. The human reference summary is additionally considered as a pseudo-model (\textbf{RefSum}) to establish an upper-bound benchmark for relational consistency.

\subsubsection{Model Output Generation}

Each model generates zero-shot summaries for the selected dataset. The generated summaries, along with input documents and reference summaries, are processed to extract structured relation triples. These triples form the basis for computing relational overlap and hallucination factors.

\subsection{Relation Extraction Framework}

We propose a dependency-aware relation extraction framework designed to enhance structural robustness and reduce spurious tuple generation. Relations are represented as normalized subject–verb–object triples extracted using syntactic dependency parsing.

The overall relation extraction is formalized in Algorithm~\ref{alg:relation_pipeline}. 
The verb-centric extraction procedure, detailed in Algorithm~\ref{alg:verb_relations}, identifies predicate-driven relational structures while incorporating multiple linguistic constraints. 
Nominal fallback extraction for prepositional noun patterns is described in Algorithm~\ref{alg:nominal_relations}, and normalization together with subject grounding mechanisms are specified in Algorithm~\ref{alg:normalization}.

The framework integrates several linguistic refinements to improve structural stability. 
Lemmatization-based normalization reduces morphological variation across predicates and arguments. 
To mitigate underspecified subject ambiguity, named entity grounded subject resolution replaces generic subjects with entity-aligned mentions when available. 
Passive constructions are addressed through agent recovery to restore implicit actors, while negation-aware verb modeling preserves relational polarity. 
Reporting verbs are filtered to exclude meta-discourse structures that do not contribute to factual content. 
Furthermore, nominal fallback extraction captures noun–preposition–object configurations, and clausal propagation enables relation recovery from embedded predicate structures. 
Finally, systematic deduplication prevents redundant triples from inflating hallucination measurements.

These refinements collectively improve structural consistency of extracted relations and reduce artificial inflation of hallucination factors. The resulting triples serve as structured inputs for computing the refined evaluation metrics, including EF1  and the normalized Relation Hallucination Index (RHI), described in subsequent sections.

\begin{figure}[t]
\centering
\includegraphics[width=0.45\textwidth]{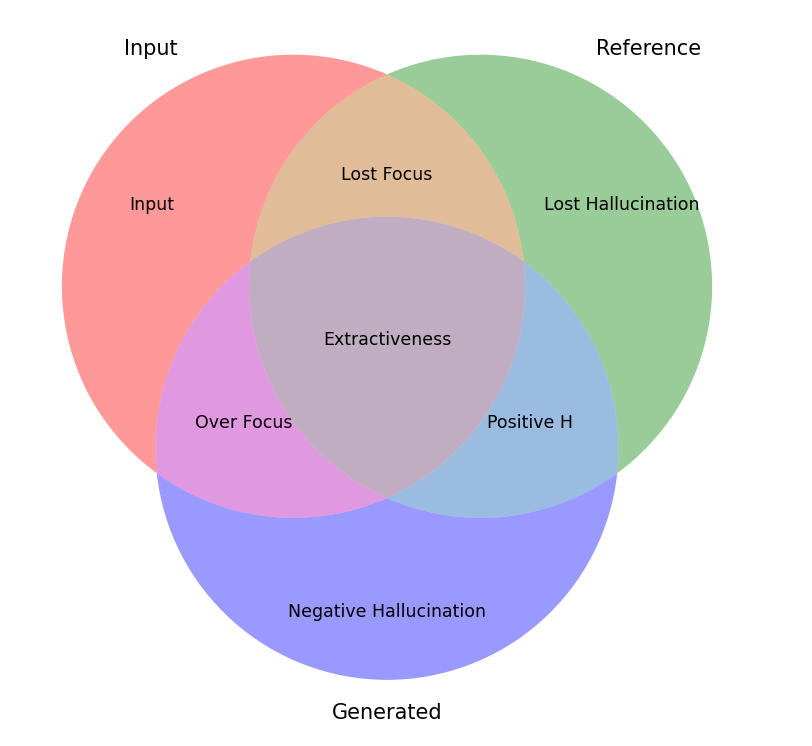}
\caption{Venn Diagram representing the six factors to identify
the relation hallucination}
\label{fig:relation_venn}
\end{figure}

\begin{algorithm}[!htbp]
\caption{Grounded Relation Extraction Algorithm}
\label{alg:relation_pipeline}
\begin{algorithmic}[1]

\Function{ExtractRelationsWithGrounding}{$text$}
\State $Relations \gets \emptyset$
\State $text \gets CleanText(text)$
\State $Chunks \gets SentenceChunks(text)$

\For{each $chunk \in Chunks$}
    \State Parse $chunk$ using dependency parser
    
    \For{each $sentence$ in $chunk$}
        \State $Relations \gets Relations \cup$
        \Statex \hspace{1.2cm} \Call{ExtractVerbRelations}{$sentence$}
        
        \State $Relations \gets Relations \cup$
        \Statex \hspace{1.2cm} \Call{ExtractNominalRelations}{$sentence$}
    \EndFor
\EndFor

\State Remove duplicate triples
\State \Return $Relations$
\EndFunction

\end{algorithmic}
\end{algorithm}

\begin{algorithm}[!htbp]
\caption{Verb-Based Relation Extraction}
\label{alg:verb_relations}
\begin{algorithmic}[1]

\Function{ExtractVerbRelations}{$sentence$}
\State $R \gets \emptyset$

\For{each $token$ in $sentence$}
    \If{$token.pos \neq$ VERB}
        \State continue
    \EndIf

    \State $v \gets Lemma(token)$
    \If{$v \in ReportingVerbs$}
        \State continue
    \EndIf

    \If{token has negation}
        \State $v_{final} \gets$ ``not\_'' $+$ $v$
    \Else
        \State $v_{final} \gets v$
    \EndIf

    \State Identify $Subjects$ from \{nsubj,nsubjpass\}
    \State Identify $Objects$ from \{dobj,obj,attr\}

    \Comment{Passive Agent Recovery}
    \For{each agent dependency}
        \State Recover subject
    \EndFor

    \For{each $s \in Subjects$}
        \If{$s.pos \notin \{NOUN,PROPN\}$}
            \State continue
        \EndIf

        \State $s_{norm} \gets$ \Call{GroundSubject}{$s,sentence$}
        \If{$s_{norm} = NULL$}
            \State continue
        \EndIf

        \For{each $o \in Objects$}
            \State $o_{norm} \gets$ \Call{Normalize}{$o$}
            \If{$o_{norm} \neq NULL$}
                \State Add $(s_{norm},v_{final},o_{norm})$ to $R$
            \EndIf
        \EndFor

        \Comment{Clausal Propagation}
        \For{each child $c$ with dep $\in \{xcomp,ccomp\}$}
            \State Extract inner verb and object
            \State Add propagated relation
        \EndFor

        \Comment{Verb Conjunction Handling}
        \For{each conjunct verb}
            \State Add conjunction relation
        \EndFor
    \EndFor

\EndFor

\State \Return $R$
\EndFunction

\end{algorithmic}
\end{algorithm}

\begin{algorithm}[!htbp]
\caption{Nominal Relation Fallback}
\label{alg:nominal_relations}
\begin{algorithmic}[1]

\Function{ExtractNominalRelations}{$sentence$}
\State $R \gets \emptyset$

\For{each noun $n$ in $sentence$}
    \For{each child with dep = prep}
        \For{each grandchild with dep = pobj}
            \State $s \gets$ \Call{Normalize}{$n$}
            \State $v \gets Lemma(prep)$
            \State $o \gets$ \Call{Normalize}{$pobj$}
            \If{$s \neq NULL$ and $o \neq NULL$}
                \State Add $(s,v,o)$ to $R$
            \EndIf
        \EndFor
    \EndFor
\EndFor

\State \Return $R$
\EndFunction

\end{algorithmic}
\end{algorithm}

\begin{algorithm}[!htbp]
\caption{Normalization and Subject Grounding}
\label{alg:normalization}
\begin{algorithmic}[1]

\Function{Normalize}{$token$}
    \If{$token$ is stopword or punctuation}
        \State \Return NULL
    \EndIf
    \If{$token.pos \notin \{NOUN,PROPN,VERB,ADJ\}$}
        \State \Return NULL
    \EndIf
    \State \Return lowercase lemma of $token$
\EndFunction

\Function{GroundSubject}{$token,sentence$}
    \If{$token$ not in generic subject list}
        \State \Return \Call{Normalize}{$token$}
    \EndIf

    \For{each named entity in $sentence$}
        \If{$entity.label \in \{PERSON,ORG,GPE,NORP\}$}
            \State \Return lowercase entity text
        \EndIf
    \EndFor

    \State \Return \Call{Normalize}{$token$}
\EndFunction

\end{algorithmic}
\end{algorithm}

\subsection{Evaluation Metrics}

To comprehensively assess factual and relational faithfulness, we employ lexical, semantic, and relation-aware evaluation metrics. Conventional measures including \textbf{ROUGE}~\cite{Lin2004ROUGE} and \textbf{SummaC}~\cite{laban2021summac} are used as comparative baselines, while relation-oriented metrics such as \textbf{EF1} and the proposed \textbf{Relation Hallucination Index (RHI)} quantify structured factual consistency. The Entity Hallucination Index (\textbf{EHI})~\cite{Praveenkumar2023EHI} is additionally considered to contrast entity-level behavior with relation-level evaluation.

\subsubsection{Lexical and Consistency Metrics}

ROUGE evaluates lexical overlap between reference ($R$) and generated summary ($S$):

\begin{equation}
\text{ROUGE-N}=
\frac{\sum_{gram_n \in R}\text{Count}_{match}(gram_n)}
{\sum_{gram_n \in R}\text{Count}(gram_n)}
\end{equation}

Relational alignment is summarized using Extractive-Faithfulness F1 (EF1)~\cite{olek2023ef1}:

\begin{equation}
EF1=\frac{2 \times Precision \times Recall}
{Precision + Recall}
\end{equation}

\[
Precision=\frac{|I \cap G|}{|G|},
\quad
Recall=\frac{|R \cap G|}{|R|}
\]

SummaC measures document--summary consistency using entailment aggregation \cite{laban2021summac}:

\begin{equation}
\text{SummaC}(D,S)=
\frac{1}{|S|}
\sum_{s_i \in S}
\max_{d_j \in D}
\text{Entail}(d_j,s_i)
\end{equation}

Coherence is computed as the average semantic similarity between adjacent sentences using sentence-level embeddings:

\begin{equation}
\text{Coherence}(S) =
\frac{1}{N-1}
\sum_{i=1}^{N-1}
\cos \left( \mathbf{e}_i , \mathbf{e}_{i+1} \right)
\end{equation}

where $S=\{s_1, s_2, ..., s_N\}$ denotes the generated summary consisting of $N$ sentences, and $\mathbf{e}_i$ represents the sentence embedding of sentence $s_i$. Higher values indicate stronger semantic continuity between consecutive sentences~\cite{Reimers2019SBERT}.
\subsubsection{Entity Hallucination Index}

Entity-level factual reliability is measured using EHI~\cite{Praveenkumar2023EHI}:

\begin{equation}
EHI=
\frac{e^{PH}+e^{EF}}
{e^{PH}+e^{EF}+e^{NH}+e^{OF}+e^{LF}}
\end{equation}

\subsubsection{Relation Hallucination Decomposition}

Let $I$, $R$, and $G$ denote relation triples extracted from input, reference, and generated summaries.

\begin{equation}
EF=\frac{3|I \cap R \cap G|}{|I|+|R|+|G|}
\end{equation}

\begin{equation}
PH=\frac{2|R \cap G|}{|R|+|G|}
\end{equation}

\begin{equation}
OF=\frac{2(|I \cap G|-|I \cap R \cap G|)}{|I|+|G|}
\end{equation}

\begin{equation}
NH=
\frac{\left||G|-(|R \cap G|+|I \cap G|-|I \cap R \cap G|)\right|}
{|G|}
\end{equation}

\begin{equation}
LF=\frac{|R|-(|I \cap R|-|I \cap R \cap G|)}
{|R|+|G|}
\end{equation}

\begin{equation}
LH=\frac{|I|-|I \cap G|}
{|I|+|G|}
\end{equation}

\subsubsection{Relation Hallucination Index}

The proposed RHI aggregates positive and hallucinated relational behaviors:

\begin{equation}
RHI=
1+\frac{EF+PH}{2}
-\frac{OF+NH+LH+LF}{4}
\label{eq:rhi}
\end{equation}

\begin{equation}
RHI_{norm}=
\frac{RHI-RHI_{min}}
{RHI_{max}-RHI_{min}}
\end{equation}

\section{Results}

We evaluate relational faithfulness using EF1, EHI, ROUGE, SummaC, coherence, and the proposed Relation Hallucination Index (RHI). Results are reported separately for abstractive news datasets (XSUM+XLSUM) and the CNN/DailyMail SumEval benchmark to analyze cross-domain consistency. 

\subsection{Average Performance on XSUM + XLSUM}

\begin{table}[!t]
\centering
\small
\caption{Average Metrics on XSUM + XLSUM}
\label{tab:xsum_metrics}
\begin{tabular}{lccccc}
\toprule
Metric & BART & PEGASUS & T5 & GPT3.5 & RefSum \\
\midrule
EF1 & 0.1230 & 0.1412 & 0.1149 & \textbf{0.1633} & 0.2885 \\
EHI & 0.6112 & \textbf{0.6162} & 0.5274 & 0.5279 & 0.9580 \\
RHI & \textbf{0.7211} & 0.6961 & 0.7012 & 0.6830 & 0.9206 \\
SummaC & 0.4312 & 0.3553 & 0.2555 & 0.0553 & 0.7838 \\
Coherence & 0.2840 & \textbf{0.8747} & 0.2812 & 0.3533 & 0.9823 \\
ROUGE-1 & 0.2140 & \textbf{0.4387} & 0.2162 & 0.2194 & 1.0000 \\
\bottomrule
\end{tabular}
\end{table}

Table~\ref{tab:xsum_metrics} shows that lexical and entailment-based metrics favor PEGASUS due to stronger surface alignment and discourse coherence. However, RHI assigns the highest score to BART, indicating improved preservation of source relations despite moderate ROUGE values. This divergence highlights RHI’s ability to detect relational correctness beyond token overlap or semantic similarity. GPT-3.5 achieves higher EF1 but lower RHI, suggesting increased relation insertion variability typical of generative models under highly abstractive settings.

\subsection{Performance on CNN/DailyMail SumEval}

\begin{table}[!t]
\centering
\small
\caption{Average Metrics on CNN/DailyMail SumEval}
\label{tab:cnn_metrics}
\begin{tabular}{lccccc}
\toprule
Metric & BART & PEGASUS & T5 & DistilBERT & RefSum \\
\midrule
EF1 & \textbf{0.1900} & 0.1600 & 0.1554 & 0.0000 & 0.7495 \\
EHI & 0.5712 & \textbf{0.5890} & 0.5454 & 0.5478 & 0.9495 \\
RHI & \textbf{0.7681} & 0.7470 & 0.7476 & 0.7188 & 0.9588 \\
ROUGE-1 & 0.3511 & 0.3419 & 0.3035 & 0.0000 & 1.0000 \\
SummaC & 0.0906 & 0.0747 & 0.0453 & \textbf{0.1570} & 0.8460 \\
RHI$_{norm}$ & \textbf{0.7912} & 0.7556 & 0.6987 & 0.5677 & 0.9851 \\
\bottomrule
\end{tabular}
\end{table}

Across the SumEval benchmark (Table~\ref{tab:cnn_metrics}), BART consistently achieves the highest RHI and EF1, demonstrating stable relational grounding on longer documents. DistilBART records near-zero ROUGE and EF1 scores, indicating failure in effective summary generation rather than metric bias. Despite moderate SummaC values, its reduced RHI confirms substantial loss of relational structure, validating the sensitivity of RHI to generation collapse scenarios.

\subsection{Statistical Validation}

Statistical analysis across both the XSUM+XLSUM and CNN/DailyMail SumEval datasets confirms the discriminative strength of the proposed RHI metric. Paired t-test results show statistically significant performance differences among summarization models ($p<0.01$), indicating that RHI consistently distinguishes summaries with well-preserved relations from those containing higher levels of hallucination. In contrast to ROUGE, EF1, EHI, and SummaC, which often yield comparable scores across models, RHI provides clearer separation and stable ranking across datasets, demonstrating its effectiveness for relation-level factual evaluation.

\subsection{Baseline vs Updated RHI Framework}

\begin{table}[!t]
\centering
\small
\caption{Baseline vs Updated RHI (XSUM+XLSUM)}
\label{tab:rhi_improve}
\begin{tabular}{lccc}
\toprule
Model & Baseline & Updated & $\Delta$ \\
\midrule
BART & 0.6378 & \textbf{0.7211} & +0.0833 \\
PEGASUS & 0.6137 & 0.6961 & +0.0824 \\
T5 & 0.6240 & 0.7012 & +0.0772 \\
GPT-3.5 & 0.6156 & 0.6830 & +0.0674 \\
\bottomrule
\end{tabular}
\end{table}
Tables~\ref{tab:rhi_improve} and~\ref{tab:ablation} show consistent RHI improvements across all models, where refinement contributes the major performance gain ($\Delta_{ref}$) and normalization provides additional stabilization ($\Delta_{norm}$). The cumulative improvements match the overall $\Delta$ observed in Table~\ref{tab:rhi_improve}, confirming the effectiveness of the proposed framework design.

\begin{table}[!t]
\centering
\small
\caption{Ablation Study of RHI Components (XSUM+XLSUM)}
\label{tab:ablation}

\resizebox{\columnwidth}{!}{
\begin{tabular}{lccccc}
\toprule
Model & RHI$_{base}$ & RHI$_{refined}$ & RHI$_{norm}$ & $\Delta_{ref}$ & $\Delta_{norm}$ \\
\midrule
BART & 0.6378 & 0.7012 & 0.7211 & +0.0634 & +0.0199 \\
PEGASUS & 0.6137 & 0.6754 & 0.6961 & +0.0617 & +0.0207 \\
T5 & 0.6240 & 0.6889 & 0.7012 & +0.0649 & +0.0123 \\
GPT-3.5 & 0.6156 & 0.6627 & 0.6830 & +0.0471 & +0.0203 \\
\bottomrule
\end{tabular}
}
\end{table}

\subsection{Cross-Dataset Hallucination Behaviour}

\begin{table}[!t]
\centering
\footnotesize
\caption{Cross-Dataset Hallucination Factors and Normalized RHI}
\label{tab:cross_dataset_hall}
\setlength{\tabcolsep}{3pt}
\begin{tabular}{lcccc}
\toprule
\textbf{XSUM + XLSUM} & BART & PEGASUS & T5 & GPT-3.5 \\
\midrule
EF  & 0.0249 & 0.0270 & 0.0213 & 0.0294 \\
PH  & 0.0635 & 0.1478 & 0.0569 & 0.0798 \\
OF  & 0.2616 & 0.0707 & 0.2089 & 0.2125 \\
NH  & 0.0873 & 0.3364 & 0.1711 & 0.4846 \\
LF  & 0.2418 & 0.2903 & 0.2410 & 0.1682 \\
LH  & 0.6907 & 0.8277 & 0.7272 & 0.6171 \\
RHI$_{norm}$ & \textbf{0.6912} & 0.6553 & \textbf{0.7055} & 0.6155 \\
\midrule
\textbf{CNN / SumEval} & BART & PEGASUS & T5 & DistilBART \\
\midrule
EF  & 0.0347 & 0.0261 & 0.0260 & 0.0000 \\
PH  & 0.1256 & 0.1154 & 0.0957 & 0.0000 \\
OF  & 0.1319 & 0.0886 & 0.1376 & 0.0000 \\
NH  & 0.1266 & 0.1902 & 0.1790 & 0.0000 \\
LF  & 0.1258 & 0.1217 & 0.1045 & 0.1249 \\
LH  & 0.8267 & 0.8701 & 0.8177 & 1.0000 \\
RHI$_{norm}$ & \textbf{0.7912} & 0.7556 & 0.6987 & 0.5677 \\
\bottomrule
\end{tabular}
\end{table}

Across both datasets, hallucination decomposition reveals consistent relational trends supporting RHI evaluation. Models with balanced Extractiveness (EF) and Positive Hallucination (PH) while maintaining reduced Over-Focus (OF) achieve higher normalized RHI scores, indicating stable relation preservation. Improvements observed on the CNN/DailyMail benchmark demonstrate that the proposed framework remains robust under longer and less abstractive summaries.

Notably, generation failure in DistilBART results in near-zero relational factors, leading to reduced RHI$_{norm}$ despite high coherence, confirming that RHI evaluates factual grounding rather than fluency. The consistent separation of models across datasets validates that aggregated hallucination factors collectively strengthen RHI’s sensitivity and enable reliable cross-domain hallucination assessment.

\subsection{Distributional Analysis of RHI}

\begin{figure}[!t]
\centering
\includegraphics[width=0.45\textwidth]{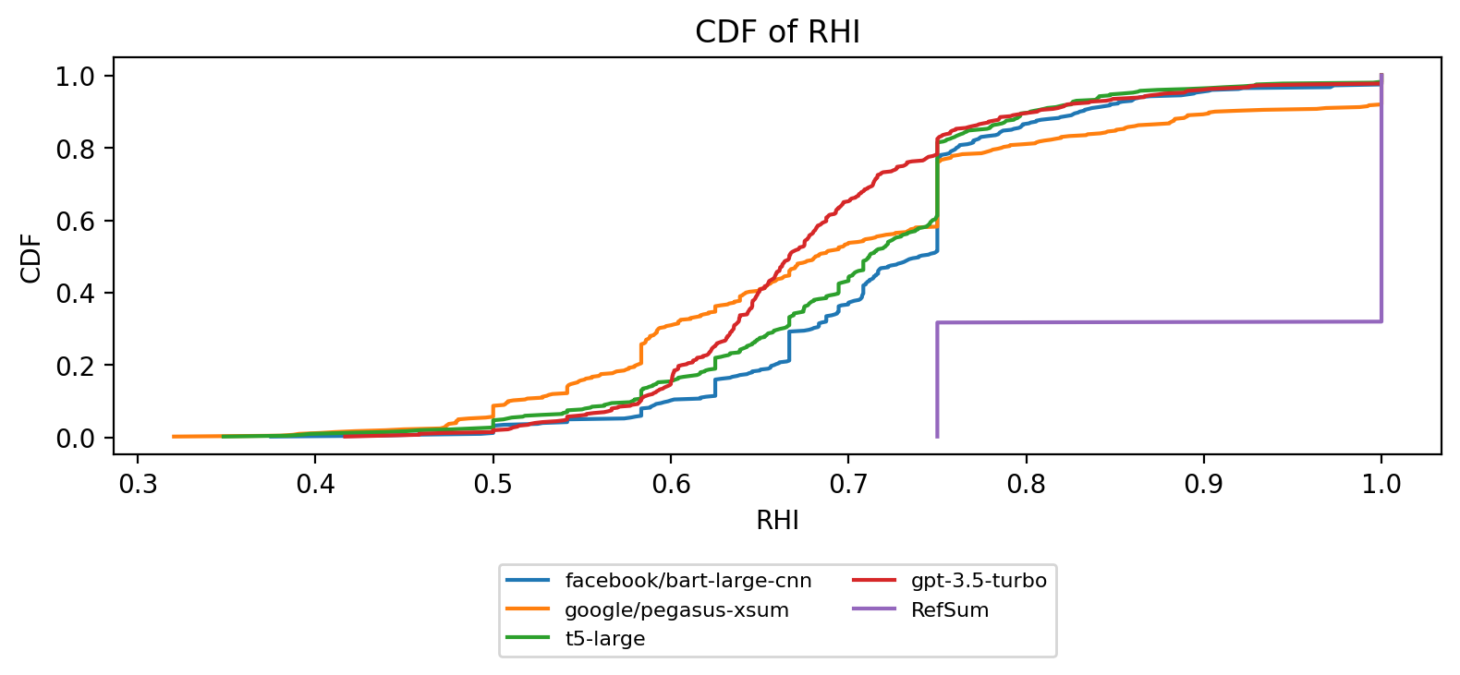}
\caption{Cumulative Distribution of RHI across models for XLSum/XSum Dataset}
\label{fig:rhi_line}
\end{figure}

\begin{figure}[!t]
\centering
\includegraphics[width=0.45\textwidth]{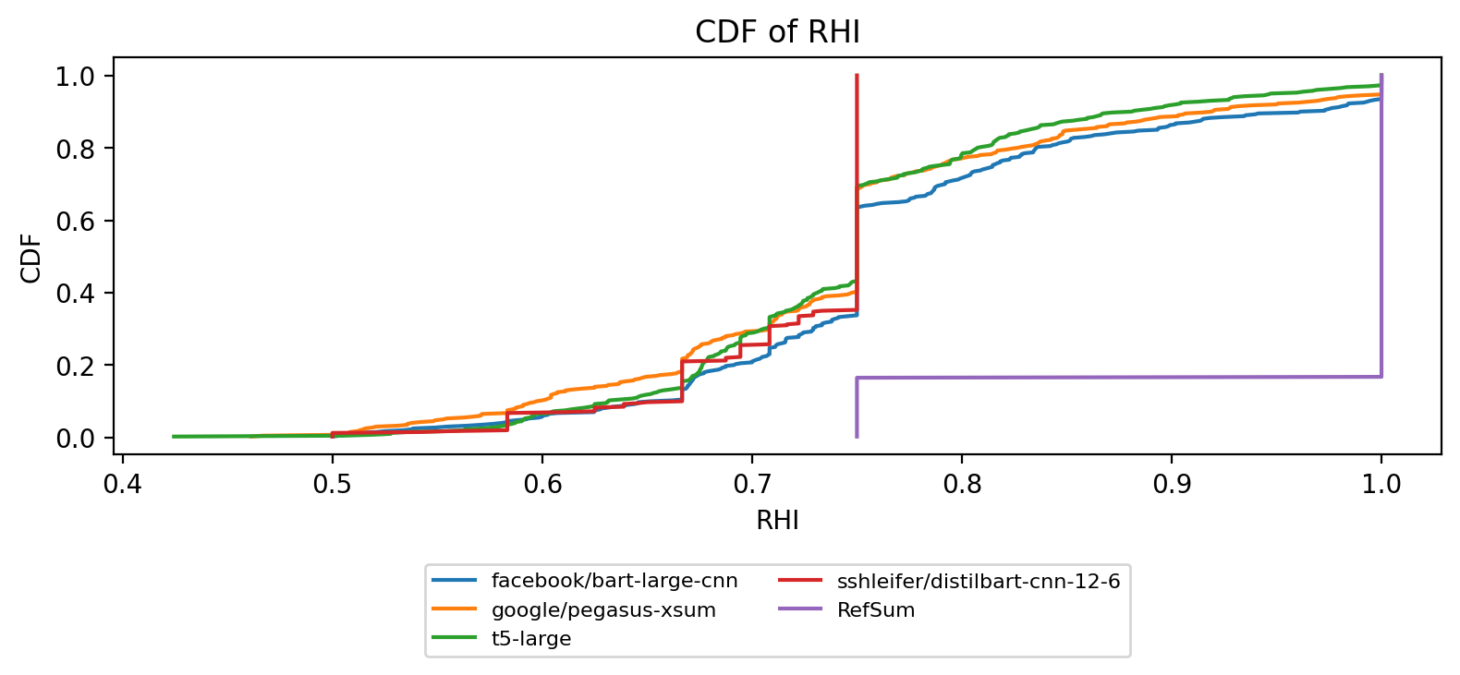}
\caption{Cumulative Distribution of RHI across models for CNN/Daily Mail Dataset}
\label{fig:rhi_cdf}
\end{figure}

Figures~\ref{fig:rhi_line} and~\ref{fig:rhi_cdf} present the cumulative distribution of RHI scores across the abstractive (XSUM+XLSUM) and long-document (CNN/DailyMail) evaluation settings. In both datasets, BART and PEGASUS exhibit smoother right-shifted distributions, indicating consistent preservation of source relations across a large proportion of summaries. T5 shows moderate dispersion, reflecting variable relational grounding under different contextual conditions. In contrast, DistilBART demonstrates an earlier saturation and compressed distribution in Fig.~\ref{fig:rhi_cdf}, revealing reduced relational coverage and confirming performance degradation observed in quantitative evaluation. The consistent ordering of model curves across Fig.~\ref{fig:rhi_line} and Fig.~\ref{fig:rhi_cdf} further indicates that RHI maintains stable discriminative behaviour independent of dataset abstraction level or document length.

\subsection{Discussion}

The experimental evaluation demonstrates that analysing summaries through relational structure provides insights that are not reflected by overlap- or entailment-oriented metrics. Across both evaluation settings, models exhibiting comparable lexical quality show noticeable differences when assessed using relation consistency, indicating that factual reliability depends strongly on preserved entity–event interactions rather than surface similarity. The grounded extraction strategy enables stable identification of meaningful relations, reducing sensitivity to stylistic variation across datasets.

Furthermore, normalized RHI scores maintain consistent model ordering under both highly abstractive and long-document summarization scenarios. The observed distributional separation confirms that relation decomposition effectively exposes hallucination patterns arising from omission, distortion, or unsupported relation generation. These findings suggest that relation-aware evaluation offers a practical mechanism for examining factual behaviour at structural granularity while remaining adaptable across model architectures and dataset characteristics.

\section{Conclusion and Future Work}

This paper introduced a grounded framework for assessing relation-level hallucination in abstractive summarization through the proposed Relation Hallucination Index (RHI). Experimental results across diverse datasets demonstrate that the proposed formulation provides stable and discriminative evaluation of factual consistency, enabling clearer differentiation between relation-preserving and hallucination-prone summarization systems. The study shows that modelling factuality at the level of entity–event interactions offers improved diagnostic capability compared to traditional evaluation measures focused primarily on lexical or semantic similarity.

Future research will investigate extending the framework toward document-level reasoning by incorporating cross-sentence relation modeling and richer contextual representations. Another promising direction involves leveraging RHI-driven feedback during model optimization to encourage generation mechanisms that prioritize factual grounding alongside linguistic fluency.
\FloatBarrier

\vspace{12pt}
\end{document}